%% file: ms.tex
\pdfoutput=1 
\documentclass[runningheads,usenames,dvipsnames]{llncs}
\usepackage[activate=true,final]{microtype}

\usepackage[hidelinks]{hyperref}

\usepackage[T1]{fontenc}
\usepackage[table]{xcolor}
\definecolor{lightgray}{gray}{1}
\usepackage{booktabs} %

\usepackage{xfrac}
\usepackage{amsmath}
\usepackage{mathtools}
\usepackage[ruled, vlined, linesnumbered]{algorithm2e}
\SetAlFnt{\small}
\SetAlCapNameFnt{\small}
\SetAlCapFnt{\small}
\SetFuncSty{textsc}

\SetCommentSty{mycommfont}

\SetKwProg{Fn}{Function}{}{}
\SetKwInOut{Input}{input}
\SetKwInOut{Output}{output}

\usepackage{caption}
\usepackage{subcaption}
\usepackage{todonotes}
\usepackage{paralist}

\usepackage{graphicx}
\usepackage{adjustbox}
\usepackage{setspace} 
\makeatletter
\def\BState{\State\hskip-\ALG@thistlm}
\makeatother

\usepackage{array}                                            %
\usepackage{longtable}                                        %
\usepackage{calc}                                             %
\usepackage{multirow}                                         %
\usepackage{hhline}                                           %
\usepackage{ifthen}                                           %

\usepackage{xparse}
\usepackage{xspace}

\newcommand{\techSEES}{SGP-DT\@\xspace}
\newcommand{\tool}{SGP-DT\@\xspace}
\newcommand{\techSEESem}{DT-EM\@\xspace}
\newcommand{\techSEESnm}{DT-NM\@\xspace}
\newcommand{\toole}{\techSEESem}
\newcommand{\tooln}{\techSEESnm}
\newcommand{\lasso}{\textsc{lasso}\@\xspace}
\newcommand{\lexicase}{$\epsilon$-\textsc{lexicase}\@\xspace}
\usepackage{breakcites}

\usepackage{multirow}
\usepackage{color, colortbl}

\usepackage{makecell}

\begin{document}
\newcommand{\boldm}[1] {\mathversion{bold}#1\mathversion{normal}}
\def\inputGnumericTable{}

\vspace{-8mm}
\title{SGP-DT: Semantic Genetic Programming \\Based on Dynamic Targets\thanks{\textbf{This is the authors version of this work. It was later published in: European Conference on Genetic Programming (EuroGP~'20)}}} %
\vspace{-4mm}
\author{Stefano Ruberto\inst{1} \and
Valerio Terragni\inst{2} \and
Jason H. Moore\inst{1}}
\authorrunning{Ruberto et al.}
\institute{Institute for Biomedical Informatics, University of Pennsylvania, United States
\email{ stefano.ruberto@pennmedicine.upenn.edu 
jhmoore@upenn.edu}\\
\and
Faculty of Informatics, Università della Svizzera italiana USI, Switzerland
\email{valerio.terragni@usi.ch}
}

\maketitle              %
\vspace{-8mm}
\begin{abstract}

	Semantic GP is a promising approach that introduces semantic awareness during genetic evolution. 
	This paper presents a new Semantic GP approach based on Dynamic Target (\tool) that divides the search problem into multiple GP runs. The evolution in each run is guided by a new (dynamic) target based on the residual errors. To obtain the final solution, \tool combines the solutions of each run using  linear scaling.
	\tool presents a new methodology to produce the offspring that does not rely on the classic crossover.
	The synergy between such a methodology and linear scaling yields to final solutions with low approximation error and computational cost.
	We evaluate \tool  on eight well-known data sets and compare with  \lexicase, a state-of-the-art evolutionary technique. \tool achieves small RMSE values, on average 23.19\% smaller than the one of \lexicase.

\end{abstract}

\input{body-conf}

\end{document}

%% file: body-conf.tex
\section{Introduction}

Recently, researchers successfully applied Semantic methods to Genetic Programming  (SGP) on different domains, showing promising results~\cite{Vanneschi2014,pawlak_review_2015,NeillSurvey2016}.
While the classic GP operators (e.g., selection, crossover and mutation) act at the syntactic level, blindly to the semantic (behavior) of the individuals (e.g., programs), the key idea of SGP is to apply semantic evaluations~\cite{Vanneschi2014}.
More specifically, classic GP operators ignore the behavioral characteristic of the offspring, 
focusing only on improving the fitness of the individuals.
Differently, SGP uses a richer feedback during the evolution that incorporates
semantic awareness, which has the potential to improve the power of genetic programming~\cite{Vanneschi2014}.

In this paper, we are considering the Symbolic Regression domain, and thus assuming the availability of training cases (defined as $m$ pairs of inputs and desired output).
Following the most popular SGP approaches~\cite{Vanneschi2014}, we intend \textit{``semantics''} as the set of output values of a program on the training cases~\cite{hutchison_semantic_2008}. 
Such an approach obtains a richer feedback during the evolution relying on the evaluation of the individuals on the training cases. More formally, the semantics of an individual $\mathcal{I}$ is a vector $\textit{sem}(\mathcal{I})=\langle y_1,y_2,\cdots,y_m \rangle$ of responses to the $m$ inputs of the training cases.
Let $\textit{sem}(\hat{y})=\langle \hat{y_1},\hat{y_2},\cdots,\hat{y_m}\rangle$ denote the semantic vector of the target (as defined in the training set), where $\hat{y_1},\hat{y_2},\cdots,\hat{y_m}$ are the desired outputs. SGP defines \textit{semantic space}~\cite{Vanneschi2014} with a metric that characterizes the distance between the semantic vectors of the individuals $\textit{sem}(\mathcal{I})$ and the target $\textit{sem}(\hat{y})$.
SGP often relies on such a distance to compute the fitness score, inducing a unimodal fitness landscape, which avoids local optima by construction~\cite{gsgp}. %

The effectiveness of SGP depends on the availability of GP operators that can move in the semantic space towards the global optimum.
An example of semantic operator is the geometric crossover proposed by Moraglio et al.~\cite{gsgp}. 
It produces an offspring with a semantic vector that lies on the line connecting the parents in the semantic space. 
Thus, it guarantees that the offspring is no worse than the worst of the parents~\cite{gsgp}.
However, such crossover operator has the major drawback of producing individuals with an exponentially increasing size (i.e., \textit{exponential bloat})~\cite{gsgp,Vanneschi2014}. 
To avoid the exponential bloat, researchers proposed variants of this operator that minimize bloating~\cite{pawlak_review_2015} but at the cost of dropping the important guarantee of non-worsening crossover operations.

In this paper, we present a new SGP approach called \textbf{\techSEES} (\textit{\textbf{S}emantic \textbf{G}enetic \textbf{P}rogramming based on \textbf{D}ynamic \textbf{T}argets}) that minimizes the exponential bloat problem and at the same time gives a bound on the worsening of the offspring.
\tool divides the search problem into multiple GP runs. Each run is guided by a different dynamic target, which \tool updates at each run based on the residual errors of the previous run. Then, \techSEES combines the results of each run into a ``\textit{optimized}'' final solution.

In a nutshell, \techSEES works as follows.
\techSEES runs the GP algorithm (see Algorithm~\ref{alg:search}) a fixed number of times ($N_{\text{ext}}$) depending on the available budget.
We call these runs \textit{external} iterations.
As opposed to the \textit{internal} iterations (i.e., generations) that the GP algorithm performs to evolve the individuals.
Each GP run performs a fixed number of internal iterations and returns a model (i.e., the best solution) that we call \textit{partial model}.
The next external iteration runs the GP algorithm with a modified training set, where \tool replaces the $m$ desired outputs $\hat{y}_i = \langle \hat{y_1},\hat{y_2},\cdots,\hat{y_m} \rangle$ with the residual errors of the partial model returned by the previous iteration. That is, the difference between 
$sem(\mathcal{I}_i)$ and $sem(\hat{y}_{i-1})$, where $\mathcal{I}_i$ is the partial model at the $i^{th}$ iteration.
Thus, at each external iteration, the fitness function evaluates differently the individuals (because the fitness functions predicates on different training sets).
As such, each partial model focuses on a different portion of the problem, the one that most influences the fitness value.
As a result, our approach leads to dynamic targets that change at each external iteration incorporating the semantic information.
\techSEES obtains the final solution after $N_{\text{ext}}$ iterations with a linear combination in the form $\sum_{i = 0}^{N_{\text{ext}}}{a_i + b_i \cdot \mathcal{I}_i}$, where $a_i$ and $b_i$ are computed with the well-known \textit{linear scaling}~\cite{keijzer2003LS}.
There is a key advantage of using linear scaling.
Keijzers showed that linear scaling gives a bound on the error of those generated individuals that are linear scaled~\cite{keijzer2003LS}.
Therefore, \techSEES entails a bound on the worsening of the offspring at each internal and external iteration.

To reduce the exponential bloat problem, \techSEES performs the internal GP iterations relying on classic mutation operators only.
It does not rely on any form of crossover, neither geometric nor classic, and thus  avoiding their fundamental limitations. Geometric crossover leads to exponential bloat and classic crossover decreases the chance to obtain a fitness improvement because it exchanges random functionalities at random points~\cite{poli_schema_1998}.
Despite the absence of crossovers, \techSEES implicitly recombines different functionalities, similarly to a geometric crossover~\cite{gsgp}. 
This is because, each partial model focuses on a different characteristic of the problem that the fitness function recognized as important (at that iteration).
This makes the search more efficient because the evolution focuses on a single characteristic at a time leaving unaltered other (already optimized) characteristics.

We evaluated our approach on eight well-known regression problems. %
We compared \techSEES with two baselines:  \lasso a least square regression technique by Efron et al.~\cite{efron2004least}; and \lexicase a state-of-the-art SGP approach by La Cava et al.~\cite{la2016epsilon}.
The results show that our approach obtains a median RMSE on 50 runs that is, on average, 51.47\% and 23.19\% smaller than the one of \lasso and \lexicase, respectively.
Moreover, \tool requires as much as 9.26$\times$ fewer tree computations than \lexicase (4.81$\times$ on average).

The remainder of this paper is organized as follows. Section~\ref{sec:methods} describes our approach.
Section~\ref{sec:related_works} discusses the related work. Section~\ref{sec:evaluation} reports our experimental evaluation and discusses the results. Section \ref{sec:conclusion} concludes the paper.

\section{Methodology}\label{sec:methods}

\input{algo/algo}
Algorithm~\ref{alg:search} overviews the \techSEES approach.
Given the values of the independent ($\overline{x}$) and dependent ($\hat{y}$) variables of the training cases, and the number of external (\textit{$\text{N}_{\textit{ext}}$}) and internal (\textit{$\text{N}_{\textit{int}}$}) iterations, it returns the final solution (\textit{finalModel}).

\tool considers tree-like individuals with the usual non-terminal symbols: $+, -, \cdot, / $(the protected division), \textit{ERC} (between -1 and 1).
In addition, \tool considers the functions \textit{Min} and \textit{Max} that returns the minimum and maximum between two numbers, respectively. The rationale of adding the two latter symbols is to inject \textit{discontinuity} to make the linear combinations more adaptable. Although also the protected division adds discontinuity in the form of asymptotes, such discontinuity often promotes overfitting~\cite{keijzer2003LS,operators}. With \textit{Min} and \textit{Max} functions, we introduce valid discontinuities alternatives that do not suffer from the limitation of the protected division.

Algorithm~\ref{alg:search} holds out a portion of the training cases for validation (lines~\mbox{\ref{algo:split-val-start}-\ref{algo:split-val-stop}}). 
\tool will use such validation sets to construct the final solution (line~\ref{algo:validation}).
Lines \ref{algo:init-target}-\ref{algo:init-models} initialize the current target with $\hat{y}$ and the lists of the best models with the empty list.
Line~\ref{algo:for-ext-start} starts the external loop, which re-assigns $\mathcal{P}$ to a fresh randomly generated population with the \textit{ramped-half-and-half} approach (function \textsc{get-random-initial-population} of Algorithm~\ref{alg:search}).
Starting every external iteration with a new population alleviates the overfitting problem. Indeed, the syntactic structures of already evolved individuals can be too complex to adapt to a new fitness landscape or to generalize on unseen data.
To further reduce overfitting and the cost of fitness evaluation, \tool generates the initial population with 
individuals with low complexity (i.e., a few nodes).

At line~\ref{algo:int-iter}, \techSEES starts the $N_{\text{int}}$ internal iterations, which resembles the classic GP but with the addition of linear scaling and the absence of crossover. %
Before line \ref{algo:fitness} computes the fitness of each individual $\mathcal{I}$ in $\mathcal{P}$, line \ref{algo:ls} performs the linear scaling of $\mathcal{I}$~\cite{keijzer2003LS}. 
Linear scaling has the advantage of transforming the semantic of individuals 
so that their potential fit with the current target is immediately given: we do not need to wait for GP to produce a partial model that reaches the same result~\cite{keijzer2003LS}.
And thus, linear scaling reduces the number of both external and internal iterations.
Fewer iterations means populations with simpler structural complexity and less computational cost.
Reducing the complexity of the solutions may reduce overfitting~\cite{rubeq}. %

Linear scaling has another important property: it gives an upper bound on the error~\cite{keijzer2003LS}.
Recall that \techSEES considers errors on dynamic targets, which change at each iteration (at the first iteration the dynamic target is $\hat{y}$). 
To exploit such a situation, we propose a fitness function based on this upper bound.
Following Keijzer~\cite{keijzer2003LS}, we compute the linear scaling of an individual $\mathcal{I}$ as follows:

\smallskip
\begin{equation}
\mathcal{I}_{ls} = a + b \cdot \mathcal{I}
\end{equation}

\vspace{-4mm}
\begin{equation}\label{eq:b_coefficient}
\text{where} \quad a=\overline{\hat{y}} -b \cdot \overline{y}  \quad \text{and} \quad b=\dfrac{\sum_{i=1}^{n} [(\hat{y}_i- \overline{\hat{y}})\cdot (y_i- \overline{y}) ]}{\sum_{i=1}^{n}[(y_i- \overline{y})^2]}
\end{equation}

\noindent We define the following fitness function of an individual $\mathcal{I}$:

\begin{equation}\label{eq:var}
\textit{fitness}(\mathcal{I}) = \sigma^2(\textit{sem}(\mathcal{I}_\text{ls}(\overline{x}))   -  \hat{y})
\end{equation}

\noindent The rationale of this function is that the Mean Square Error (MSE) of  $\mathcal{I}_{ls}$ has the variance ($\sigma^2$) of the current target as an upper bound~\cite{keijzer_scaled_2004}:

\begin{equation} \label{eq:errobound}
\textit{MSE} = \dfrac{ \sum_{i=0}^{m} (y_i - \hat{y}_i)^2 }{m} \leq \sigma^2(\hat{y})
\end{equation}
where $m$ is the number of training cases ($y$).

\noindent At each new external iteration the residual error becomes the new target (line~\ref{algo:new-target}).
\begin{equation} \label{eq:residula-error}
\textit{target}= \hat{y} - \textit{sem}(\mathcal{I}^\star_{\textit{ls}}(\overline{x}))
\end{equation}
where $\textit{sem}(\mathcal{I}^\star_{\textit{ls}}(\overline{x}))$ is the evaluation of the best individual at the current iteration, which we call \emph{partial model}.

The inequality~\ref{eq:errobound} does not guarantee that the external iterations converge to a lower MSE because we do not know if $\sigma^2(\textit{error}) \leq \sigma^2(\hat{y})$, where  $\textit{error} = target - \textit{sem}(\mathcal{I}^\star_{\textit{les}}(\overline{x}))$.
Thus, by optimizing the variance of the error shown in equation \ref{eq:var}, we act directly on the minimization of the upper bound, so that the next external iteration can benefit from a lower bound.

At lines \ref{algo:offspring-start}-\ref{algo:offspring-stop}, Algorithm~\ref{alg:search} runs a classic GP algorithm without crossovers, using only mutations.
We use a tree-based mutation operator because \techSEES uses trees as syntactic structures for the individuals.
The operator randomly generates a subtree from a randomly chosen node.
To increase the synergy with linear scaling, we set two constraints during mutation.
First, the node selection is biased towards the leaves of the tree, so that the mutated tree does not diverge too much from the original semantic (\textit{locality principle}). 
Producing a mutation that is close to the original semantic of the tree preserves 
the validity of the selection performed after the linear scaling.
And thus, we only allow minor changes to improve the fitness.
Second, for the same reason, the mutation is biased towards replacing the selected node with a sub-tree of limited depth. 
Note that, we decided not to limit the maximum size (number of nodes in the tree) or depth of an individual.
By doing so, GP can grow and choose the right solution complexity for the problem at hand.
These two constraints help us to mitigate the overfitting and bloat problem without preventing the \techSEES to effectively search for competitive individuals.
As linear scaling helps GP to find useful individuals (thanks to the upper bound). 
Moreover, additional external iterations will further refine other aspects of the problem not yet addressed.

We decided to exclude the classic crossover operator in the internal iterations, as several researchers argued about the effectiveness of crossover in relation to the problem of modularity of GP~\cite{gerules_survey_2016}.
There is a consensus that an effective GP algorithm needs a crossover that preserves the semantics of the parts swapped among individuals respecting the boundaries of a useful functionality within the individual's structure~\cite{poli_schema_1998,pawlak_review_2015,krawiec_locally_2013}.
According to McPhee et al.~\cite{hutchison_semantic_2008} and Ruberto et al.~\cite{rubeq} most classic crossover operators do not obtain a meaningful variation (or any variation at all) in the program semantics, when dealing with Boolean and real value symbolic regression domains.
The main issue is that classic crossover operators do not preserve a \textit{common context}~\cite{hutchison_semantic_2008} among the building blocks of the individuals exchanged during crossover, which is important to increase the chance of obtaining a semantically meaningful offspring~\cite{krawiec_locally_2013}. The idea of determining a common context has been introduced by Poli and Langdon with the one-point crossover operator~\cite{poli_schema_1998}. 
But how to identify a meaningful common context among trees structures %
is still an open problem.

Instead, \techSEES  exchanges functionalities among individuals by relying on the linear combination of the \textit{partial models} (i.e., the fittest individuals at each external iteration, line~\ref{algo:best-ind} Algorithm~\ref{alg:search})
and on a specific mechanism for selecting and mutating the individuals during the GP runs.
In light of this, we exclude the crossover operators in the presence of these semantic recombination alternatives. %
To have an effective exchange of functionalities among individuals we need to: (i) preserve building blocks semantics (ii) preserve the context of building blocks (iii) make the exchange of functionalities directed towards producing new and interesting semantics. 
\techSEES achieves these objectives by (i) mapping each building block to a single partial model (this would avoid arbitrary fragmentations of the blocks); (ii) preserving the context of the building blocks because in our scenario the partial models obtained at previous iterations represent the context; and (iii) using mutation only, which promotes diversity in the population.
Despite the absence of crossover, \techSEES exchanges building blocks because each partial model is a building block.
Differently from the classical crossover that exchanges random fragments, \mbox{\techSEES} obtains the final model by summing the linear scaled partial models.
This approach makes the exchange of functionalities more effective, as each partial model (building block) characterizes a specif functionality.

The for-loop at line~\ref{algo:for-ext-start} terminates when \tool concludes all external iterations.
We decide not to introduce a different stopping criterion based on the stagnation of fitness improvement. This is because it is difficult to predict if the fitness will not escape stagnation in future iterations. After all the external iterations, the function \textsc{validate-and-select} at line~\ref{algo:validation} of Algorithm~\ref{alg:search} returns the partial models that will be combined into the final solution.
Such models are selected as follows.
The validation takes in input the ordered sequence of best individuals (\textit{models}) collected after each internal iteration (line~\ref{algo:collect-models} Algorithm~\ref{alg:search}) and the validation sets ($\overline{x}_\textit{val}$ and $\hat{y}_\textit{val}$) obtained at line \ref{algo:split-val-start}.
Note that, \tool saves the computed linear scaling parameters ($a$ and $b$ equations (\ref{eq:b_coefficient})) at line \ref{algo:ls} and do not recompute them during the validation and test phases.
Internally, the validation scans the sequence \textit{models} and progressively computes the MSE evaluating the individuals on the validation set to find the point in the sequence where MSE is the smallest.
\tool finds the smallest MSE using the rolling mean of the validation set error at a fixed window size to minimize the short-term fluctuations.
The function \textsc{validate-and-select} returns the sequence (\emph{bestModels}) of the partial models that were produced before the smallest MSE. Such sequence represents the transformation chain of the dynamic targets.
In case \tool obtained the model with the smallest MSE during the internal iterations, it appends this individual at the end of  \emph{bestModels}.
Line 
\ref{algo:linear-comb} of Algorithm~\ref{alg:search} computes the final model by summing all the models in \emph{bestModels}.

\section{Related Work}\label{sec:related_works}
This section divides the related work of \techSEES in three groups. Each group refers to techniques that are relevant to a main characteristic of \techSEES: (i) having dynamic or semantic objectives, (ii) using linear combinations or geometric operators, (iii) using an iterative approach on residual errors.

\smallskip
\noindent \textbf{Dynamic or semantic objectives~}
The GP techniques proposed by Krawiec et al.~\cite{kw15automatic} and Liskowski et al.~\cite{DSOkw2017} present semantic approaches that consider interactions between individuals %
and the training set.
These approaches cluster such interactions to derive new targets for a multi-objective GP. 

Otero et al. proposed an approach with dynamic objectives that combines intermediate solutions in a final Boolean tree~\cite{autoDecompositionBoolean2013}. This technique progressively eliminates from the training cases the ones perfectly predicted from the current intermediate solution and operates exclusively in a Boolean domain.

Krawiec and O'Reilly~\cite{kw14behavioral} proposed a GP approach that explicitly models the semantic behavior of a solution during the computation of training cases.

BPGP by Krawiec and O'Reilly~\cite{kw14behavioral} explicitly models the semantic behavior of a solution during the computation of training cases.
BPGP proposes an operator that mutates an individual by replacing a randomly selected sub-tree with a random one. %
According to Krawiec and O'Reilly this ``mutation-like''~\cite{kw14behavioral} operator is intended as a ``form of crossover''. 
We think that this is similar in principle to our design choice of dropping  crossover altogether and instead choosing among mutated alternatives in the population.
However, Krawiec and O'Reilly still use the traditional crossover alongside with this new mutation~\cite{kw14behavioral}. 

We differ from all of these techniques because we build our solution progressively crystallizing the intermediate achievements. Most of these approaches use auxiliary objectives during their search and use a single GP run. %
Conversely,  \techSEES uses a non-predetermined number of objectives in subsequent GP runs. 
The approach of Otero et al.~\cite{autoDecompositionBoolean2013}  is the only one that  progressively builds the solution %
but it uses a strategy that works for Boolean trees only.

\smallskip
\noindent \textbf{Linear combinations~}
MRGP~\cite{mrgp_2014}  uses multiple linear regression to combine the semantics of sub-programs (subtrees) to form the semantic of an individual.

Ruberto et al. proposed ESAGP~\cite{esagp}, which derives the target semantics by relying on a specific linear combination between two ``optimally aligned'' individuals in the error space.
Leveraging such geometric alignment property, Vanneschi et al. proposed  \textsc{NA-GP}~\cite{VANNESCHIESAGP}, which performs  linear combinations  between two aligned chromosomes belonging to the same individual.

Gandomi et al. proposed MGGP~\cite{Gandomi2012}, where each individual is composed of multiple trees. %
MGGP produces the final solution with a linear combination of the tree's semantics, deriving the values of the coefficients from the training data with a classic least squares method.
However, the number of trees in the linear combination is fixed and the fitness landscape is not dynamic.

Moraglio~et~al. proposed the Geometric Semantic GP (GSGP) crossover operator~\cite{gsgp}, which uses linear combinations to guarantee offspring that is not worse than the worst of the parents.
Unfortunately, GSGP suffers from the exponential bloat problem and requires many generations to converge, especially if the target is not in the convex hull spanned by the initial population~\cite{gsgp}.

Notably, all the approaches described in this second group use a single run to search for the final solution. 
Differently from \techSEES, they fix the number of components in advance (the only exception is GSGP but it suffers from the exponential bloat problem~\cite{gsgp}).
In addition, all of the techniques in the first and second groups have a static target, and thus they continuously evolve a population without re-initialization.
This limits the diversity of the genetic alternatives when the population converges at later generations. %
Conversely, \tool has a dynamic target and it starts with a fresh population at each internal iteration (see Algorithm~\ref{alg:search}).

\smallskip
\noindent \textbf{Iterative approaches  based on residual errors~}
Sequential Symbolic Regression (SSR)~\cite{oliveira2015sequential} uses the crossover operator  GSGP~\cite{gsgp} to iteratively transform the target using 
a semantic distance that resembles the classical residual approach.
However, no statistical difference (on the errors) from the classical GP approach was found~\cite{oliveira2015sequential}. 
Differently from \tool, SSR considers residuals  %
that do not optimize the linear combinations with a least square method. Although SSR overcomes the exponential bloat, it weakens the advantage of using residuals.

Medernach et al. presented the \textsc{wave} technique~\cite{wave1,wave2} that similarly to \tool, executes multiple GP runs using the same definition of residual errors (equation~\ref{eq:residula-error}) and obtains the final model by summing the intermediate models. %
\textsc{wave} produces a sequence of short and heterogeneous GP runs, obtained by ``fuzzing'' the settings of system parameters (e.g, population size, number of internal iterations ) and by alternating the use of linear scaling.
However, \mbox{\tool} drastically differs from \textsc{wave}.
The heterogeneity nature of \textsc{wave} emulates this dynamic evolutionary environment by simulating periods of a rapid change~\cite{wave1,wave2}. The effectiveness of such an approach requires specif combinations of system parameters that converges to a fitter solution. Due to the huge space of possible system parameters, finding such combinations often requires a large number of iterations~\cite{wave1,wave2}. 
Conversely, \tool steers the evolution with a novel approach that gradually evolves the building blocks of the final solution without exploring the huge space of possible combinations of system parameters.

All the techniques of this group use residuals differently from \tool. Moreover, they rely on the classic or geometric crossover.
Conversely, one of the key novel aspects of \tool is to avoid crossover altogether.

\begin{table}[t]
	\vspace{-1mm}
	\caption{Data sets of regression problems.}\label{table:dataset}
	\input{tables/subjects-new}
	\vspace{4mm}
\end{table}

\section{Evaluation} \label{sec:evaluation}

\textbf{Data sets~} 
We performed our experiments on eight well-known data sets of regression problems that have been used to evaluate most of the techniques discussed in Section~\ref{sec:related_works}~\cite{la2016epsilon,mrgp_2014,VANNESCHIESAGP,Gandomi2012,wave1,wave2}.
Table~\ref{table:dataset} shows the name,  number of attributes, and number of instances for each data set. %
For \textit{uball5d}\footnote{$f(x)=10/(5 + \sum_{i=1}^{5} (x_i -3)^2)$} we followed the same configuration used by Cava et al.~\cite{cava2018probabilistic}.

\subsection{Methods}
 We compared \techSEES with two techniques
(\lasso~\cite{efron2004least} and \lexicase~\cite{la2016epsilon}) and two variants of \techSEES (\techSEESem and \techSEESnm).

\smallskip
\noindent \textbf{\lasso~}
Both \techSEES and \lasso~\cite{efron2004least} use the least square regression method to linearly combine solution components.
More specifically, \lasso incorporates a regularization penalty into least-squares regression using an  $ \ell_1$ norm of the model coefficients and uses a tuning parameter $\lambda$ to specify the weight of this regularization~\cite{efron2004least}.  We relied on the \lasso implementation by Efron et al.~\cite{efron2004least}, which automatically chooses $\lambda$ using cross-validation.

\smallskip
\noindent \textbf{\lexicase~} 
This evolutionary technique adapts the \textit{lexicase} selection operator for continuous domains~\cite{la2016epsilon}.
The idea behind \lexicase selection is to promote candidate solutions that perform well on unique subsets of samples in the training set, and thereby maintain and promote diverse building blocks of solutions~\cite{la2016epsilon}. 
Each parent selection begins with a randomized ordering of both the training cases and the solutions in the selection pool (i.e., population). Individuals are iteratively removed from the selection pool if they are not within a small threshold ($\epsilon$) of the best performance among the pool on the current training sample. The selection procedure terminates when all but one individual is left in the pool, or until all individuals have tied performance.
In the latter case, a random one is chosen. 
The recent study of Orzechowski et al. shows that  \lexicase~\cite{la2016epsilon} outperforms many GP-inspired algorithms~\cite{bigcomparison}.
We relied on the publicly available implementation of \lexicase, \textit{ellyn}\footnote{https://github.com/EpistasisLab/ellyn}, which uses stochastic hill climbing   to tune the scalar values of each generated individual.
It also relies on a  25\% validation hold-out from the training data to choose the final model from a bi-dimensional \textit{Pareto archive}, which \textit{ellyn} constantly updates during the evolution.
The two dimensions are the number of nodes and the fitness.

\smallskip
\noindent \textbf{\techSEESem~}
We considered a variant of \techSEES (called \techSEESem) with a modified fitness function as the only difference with \techSEES:
\begin{equation} \label{eq:MSE}
\textit{fitness}(\mathcal{I})  = \textit{MSE} = \dfrac{ \sum_{i=0}^{m} (y_i - \hat{y}_i)^2 }{m}
\end{equation}

While the original fitness of \techSEES minimizes the upper bound of the MSE in equation~\ref{eq:var}, this function directly minimizes the MSE in equation \ref{eq:MSE}. This variant helps to evaluate the impact of a direct error minimization with respect to a more qualitative and indirect measure of the error, such as the variance ($\sigma^2$).  

\smallskip
\noindent \textbf{\techSEESnm~}
We considered another variant, called \techSEESnm, that excludes the \textit{Min} and \textit{Max} non-terminal symbols (as the only difference with \tool), and thus evaluating the advantage of different discontinuity types during the evolution.

\vspace{-1mm}
\subsection{Evaluation setup}

Following the setup of Orzechowski et al.~\cite{bigcomparison} for \lexicase, we set for all the four GP techniques (\techSEES, \lexicase, \techSEESem, and \techSEESnm) a population size of 1,000 and a budget of 1,000 generations.
We ran 50 trials for every technique on each data set using 25\% of the data for testing and 75\% for training.

\techSEES and its two variants share the same configuration:
We divided the 1,000 generations in 20 external iterations ($N_{\text{ext}}= 20$), and thus the number of internal iterations ($N_{\text{int}}$) is 50.
We used ramped half\&half initialization up to a maximum depth of four (function \textsc{get-random-initial-population} at line~\ref{algo:random-pop} of Algorithm~\ref{alg:search}).
The probability of mutation is 100\% and the maximum depth of the sub-trees generated by the mutation operators is five. 
The probability of a sub-tree mutation happening at the leaf level 
is 70\%. 
We set no limits on the number of nodes in the trees and on the depth of the trees.
We set the Elitism to keep only the best individual at each internal iteration (function \textsc{elite} at line~\ref{algo:elite} of Algorithm~\ref{alg:search}).
We obtained the validation set by extracting  10\% of the training cases (function \textsc{split} at line~\ref{algo:split-val-start} of Algorithm~\ref{alg:search}).
The fixed window size for the rolling-mean is 20.
We chose this configuration after a preliminary tuning phase and kept uniform for all the eight data sets.

\subsection{Results and discussion}
\smallskip
\noindent \textbf{Errors' Comparison~}
Following previous work we use the Root Mean Square Error (RMSE) to evaluate the final solution with the test set.
The first five columns of Table~\ref{table:rmse} show for each technique the median RMSE of the 50 trials.
The last four columns of Table~\ref{table:rmse} indicate the percentage decrease of the RMSE medians with respect to the competitor techniques\footnote{calculated with $((M_T- M_D)/M_T) \cdot 100$, where $M_D$ is the median RMSE of \tool and $M_T$ is the one of the competing technique}. A positive percentage value means that the RMSE median of \tool is lower (i.e., better), while a negative value means a worst median RMSE.
Figure~\ref{fig:boxTestRMSEAll} shows the box plots of the RMSE values of the 50 trials\footnote{for readability reasons we omitted 4 out-layers for \lasso, 13 for \lexicase, 30 for \tool, 30 for \tooln and 35 for \toole}.
When comparing the RMSE values we performed a non-parametric pairwise Wilcoxon rank-sum test with Holm correction for multiple-testing, with a confidence level of 95\% (p-value <0.05).

\begin{table}[t]
	\caption{Median RMSE of the 50 trials.}\label{table:rmse}
	\input{tables/rmse-new}
	
\end{table}

\tool achieves a smaller RMSE than \lasso for all the data sets, obtaining always statistical significance.
The decrease of the RMSE medians ranges from 9.06\% for \emph{housing} to 88.67\% for \emph{yacht} (51.47\% on average).
\tool has smaller RMSE medians than \lexicase for all data sets but \emph{housing} (decrease -4.48\%).
This is the only comparison of \tool and \lexicase without statistically significance. The decrease of the RMSE medians ranges from -4.48\% for \emph{housing} to 57.07\% for \emph{ench} (23.19\% on average). This is a remarkable result considering that  \lexicase outperforms many GP-inspired algorithms~\cite{bigcomparison}.
Comparing with the variant \techSEESem, \tool achieves the only statistically significant differences with \techSEESem on the data sets \textit{uball5d} and \textit{yacht}, with percentage decreases of 6.63\% and 20.45\%, respectively.
For such datasets \tool performs better than  \techSEESem indicating that our fitness function that  minimizes the upper bound achieves a better final solution.
\tool has statistically significant differences of the median RMSE  with \tooln only with the data sets \textit{airfoil}, \textit{tower} and \textit{uball5d}.
\techSEES performs better than \techSEESnm on the \textit{airfoil} and \textit{tower} datasets: 3.94\% and 10.12\% of percentage decrease, respectively. 
This means that the \textit{Min} and \textit{Max} non-terminal symbols provide an advantage only in these two datasets.
However, Figure~\ref{fig:boxTestRMSEAll}  indicates that using such non-terminal symbols does not penalize the outcome in any other dataset, except for \textit{uball5d} where the difference is statistically significant (the decrease is -7.87\%).

	\input{images/rmse-plot-new}
\smallskip
\noindent 
\textbf{Error comparison with related work~}
Unfortunately, the implementation of \textsc{wave}~\cite{wave1,wave2} is not publicly available, and thus a direct comparison would be difficult.
We extracted the median RMSE from the GECCO 2016 paper~\cite{wave2} for our two common subjects: $4.1$ (\textit{concrete}) and $8.7$ (\textit{yacht}). \tool achieves a median RMSE percentage decrease of 25.17\% (\textit{concrete}) and 75.12\% (\textit{yacht}), see Table~\ref{table:rmse} for the reference values.
Note that, the computational cost reported in the GECCO paper has the same order of magnitude with the one of \tool.

From the paper of Vanneschi et al.~\cite{VANNESCHIESAGP}, we extracted the median RMSE on the data set \emph{concrete} of the following GP techniques: 10.44 (\textsc{NA-GP}~\cite{VANNESCHIESAGP}), 8.1
(\textsc{NA-GP-50}~\cite{VANNESCHIESAGP}), 12.50 (\textsc{GSGP}~\cite{gsgp}), and 9.43 (\textsc{GSGP-LS}~\cite{castelli:truji}). \tool has a percentage decrease of 37.64\%, 19.62\%, 47.92\% and 30.96\%, respectively. These results are only indicative because their evaluation setup differs from ours.

\smallskip
\noindent \textbf{Computational effort~}
To evaluate the computational effort of the evolutionary techniques
we decided not to rely on execution time because it depends on implementation details.
Instead, we relied on the total number of evaluated nodes (being not a GP technique this metric is not applicable to \lasso).
Both \tool and \lexicase operate on nodes, \tool on tree-like data structures, while \lexicase on stack-based ones.
Following Ruberto et al.~\cite{rubeq}, we count a node operation every time a technique evaluates a node regardless the purpose of the operation (e.g., mutation, fitness computation).
We excluded the computational effort of linear scaling because it does not perform operations on nodes.
However, it has a linear computational cost of $\mathcal{O}(m\cdot P)$, where $m$ is the size of the training set and $P$ the population size. 
For comparing the number of evaluated nodes, we used the Wilcoxon rank-sum test with Holm correction for multiple-testing, with a confidence level of 95\% (p-value <0.05).
The test show that all the comparisons between each pair of techniques are statically significance, except the comparison with \tool and \tooln on subject \emph{uball5d}.

Table~\ref{table:compEffort} reports the median number of nodes (of the 50 runs) that the GP techniques evaluate to produce the final solution.
The last three columns of Table~\ref{table:compEffort} report the ratio between the number of node evaluations of \tool with those of  \lexicase, \toole and \tooln.
A ratio greater (lower) than one means that \tool evaluates a lower (higher) number of nodes.
Comparing with \lexicase, \techSEES reduces the amount of node evaluations by a factor between 4.01$\times$ and 9.26$\times$, obtaining statistically significant better RMSE values than \lexicase for seven out of eight data sets.
This result can be explained by (i) \techSEES computes only a fraction of the entire solution (partial models) at a time; (ii) the size of the individuals is kept at minimum (see Section~\ref{sec:methods}).

The number of evaluated nodes of \tool and \toole are almost identical (0.99$\times$ on average).
This indicates that guiding the evolution with the 
fitness function of \tool and with the one of \toole yield to the same computational cost but \tool achieves better median RMSE (5.39\% on average). 
\tooln always evaluated less nodes than \tool (0.77$\times$ on average).

\begin{table}[t] 
	\caption{Median number of evaluated nodes and reduction ratio of \techSEES.}\label{table:compEffort}
	\input{tables/nodes-new}
	
\end{table}

\noindent
\textbf{Size of the final solutions~}  \tool has no limits on the maximum complexity of the individuals, while \lexicase has a limit of 50 nodes because at higher limits the computational effort of \lexicase becomes prohibitively expensive~\cite{la2016epsilon}.
\tool produces solutions with size ranging from 442 to 1,184 nodes (760 on average), which is on average 15$\times$ larger than the one produced by \lexicase and is not large enough to be considered (exponential) bloat.
This extra complexity of the final solutions positively contributes at the performance of the algorithm.
We are investigating a post-processing phase to simplify the final solutions. 

On average, \toole produces solutions with 806 nodes and \tooln with 591.
\tooln generates smaller solutions than \toole, this could be due to the fact that \tooln has a smaller search space (\tooln omits the \textit{Min} and \textit{Max} symbols).
Evaluating smaller solutions require less computation, this explains why \tooln requires less computation than \tool and \toole (see Table~\ref{table:compEffort}).

\noindent \textbf{Overfitting~}
Figure~\ref{fig:boxTestMediaEAll} plots for each data set the median of the best RMSE by computational effort (number of evaluated tree nodes) for \tool and its two variants. Unfortunately, the implementation of \lexicase that we used does not report the intermediate RMSE on test.
We use the computational effort, rather the number of generations, for a fair comparison of the three techniques. This is because the number of evaluated nodes is not uniform across the generations.

\newcommand{\captionSpaceMedian}{-15pt}
\newcommand{\zfMedian}{0.30} %
\newcommand{\hspaceMedian}{0px}
\newcommand{\hspacenegMedian}{0px}
\newcommand{\spacev}{0px}
\begin{figure*}[t]

	\centering
	\vspace{\spacev}
	\hspace{\hspacenegMedian}	
	\begin{subfigure}[t]{\zfMedian\linewidth}
		\centering
		\includegraphics[width=\linewidth]{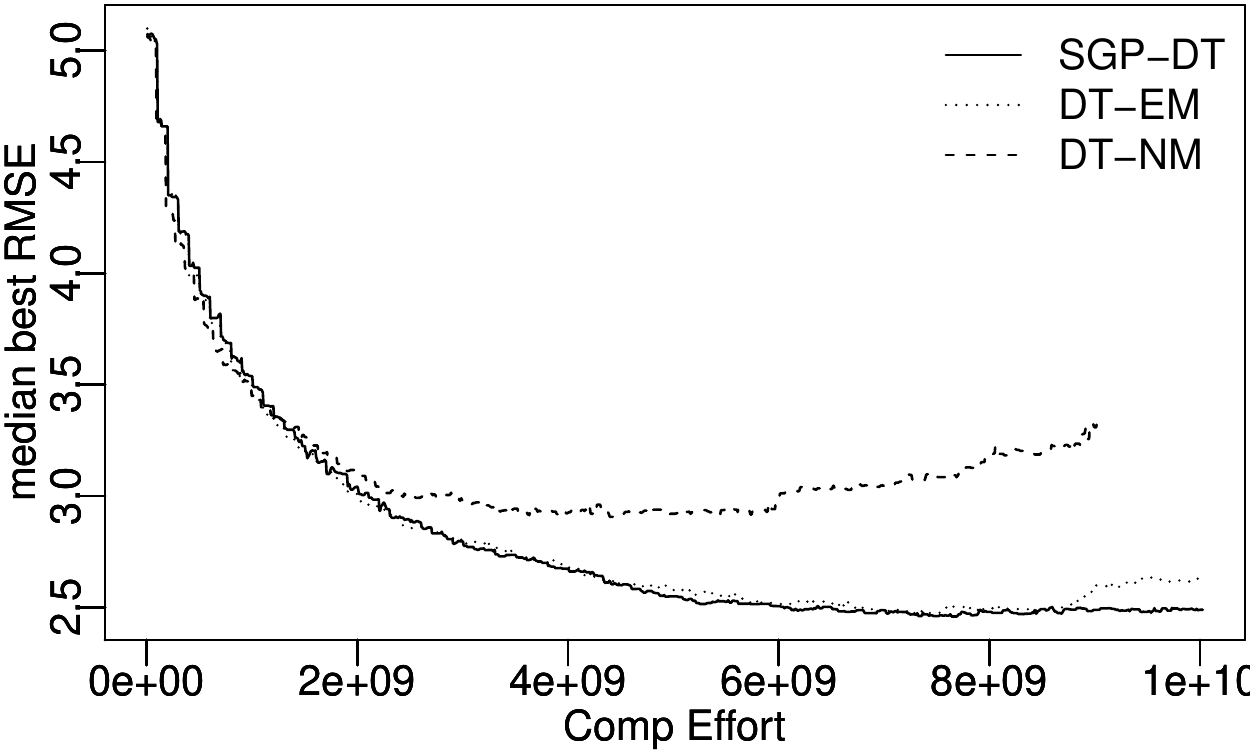}
		\vspace{\captionSpaceMedian}
		\caption{ \textbf{airfoil}}\label{fig:RMSEmedian_airfoil}		
	\end{subfigure}
\vspace{\spacev}
	\hspace{\hspaceMedian}
	\begin{subfigure}[t]{\zfMedian\linewidth}
		\centering
		\includegraphics[width=\linewidth]{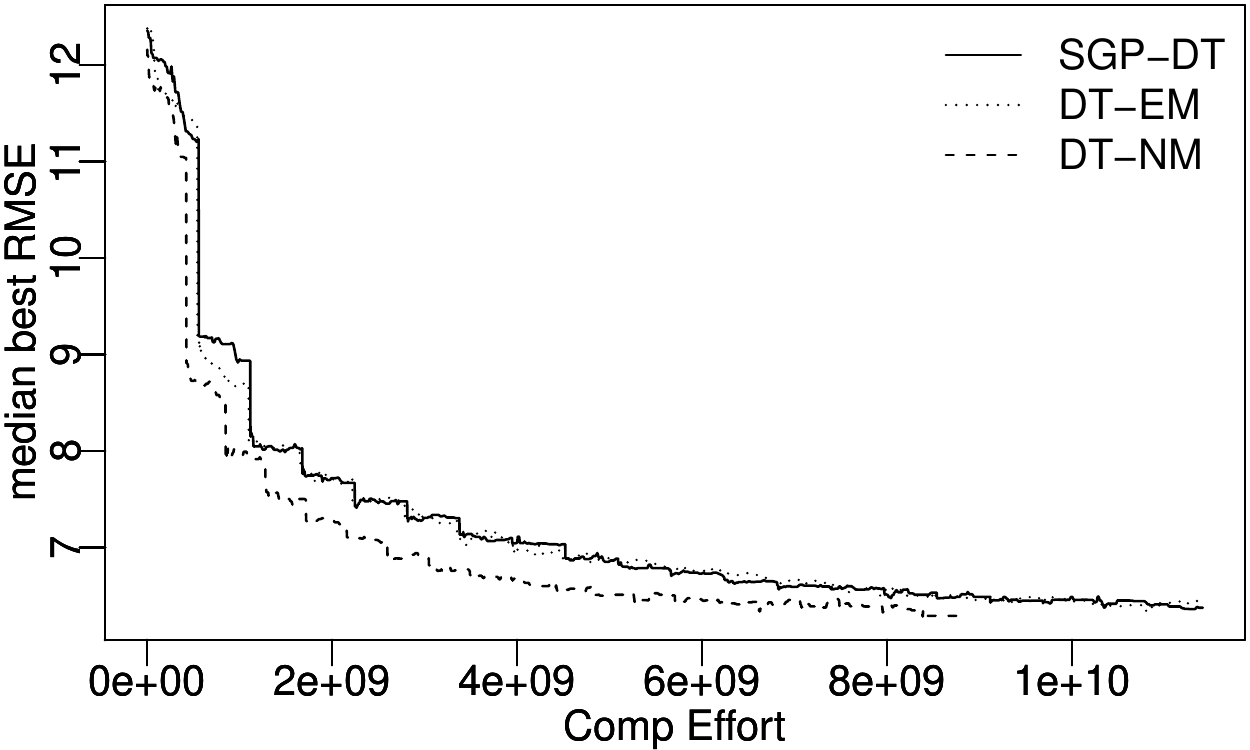}
		\vspace{\captionSpaceMedian}
		\caption{\textbf{concrete}}\label{fig:RMSEmedian_concrete}		
	\end{subfigure}
\vspace{\spacev}
	\hspace{\hspaceMedian}
	\begin{subfigure}[t]{\zfMedian\linewidth}
		\centering
		\includegraphics[width=\linewidth]{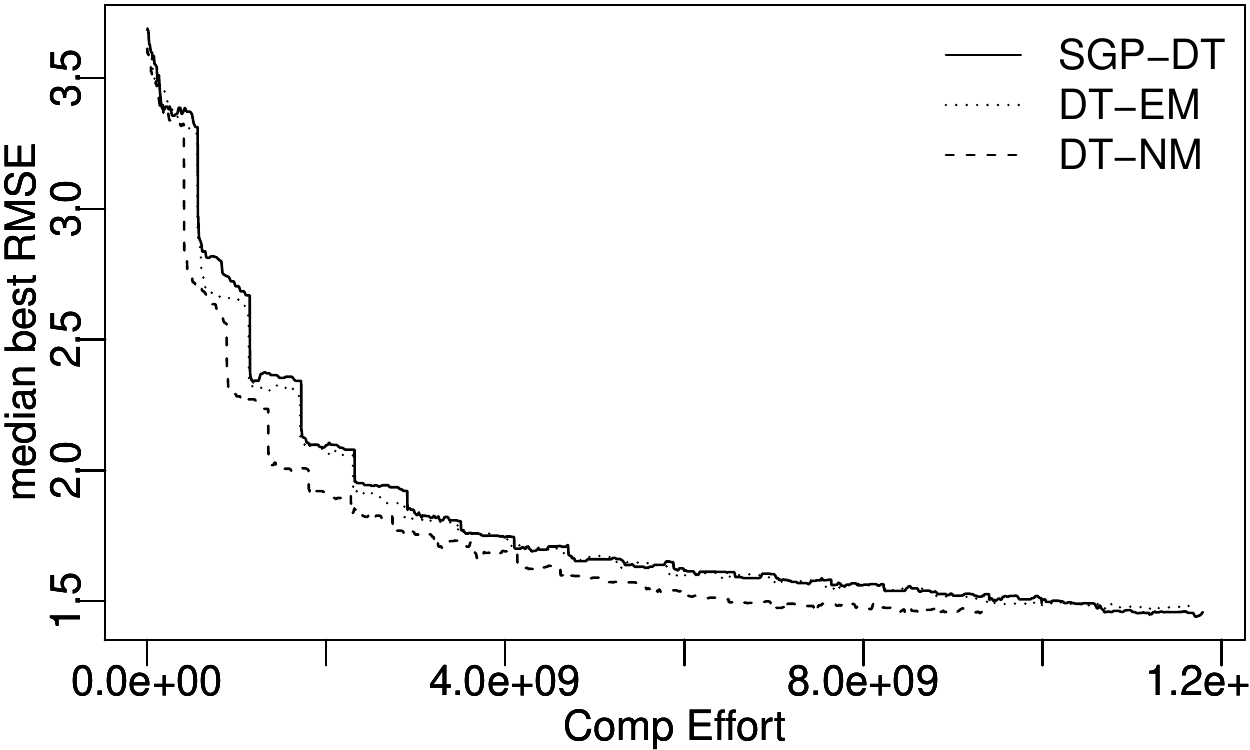}
		\vspace{\captionSpaceMedian}
		\caption{\textbf{enc}}\label{fig:RMSEmedian_enc}		
	\end{subfigure}
\vspace{\spacev}
	\hspace{\hspaceMedian}	
	\begin{subfigure}[t]{\zfMedian\linewidth}
		\centering
		\includegraphics[width=\linewidth]{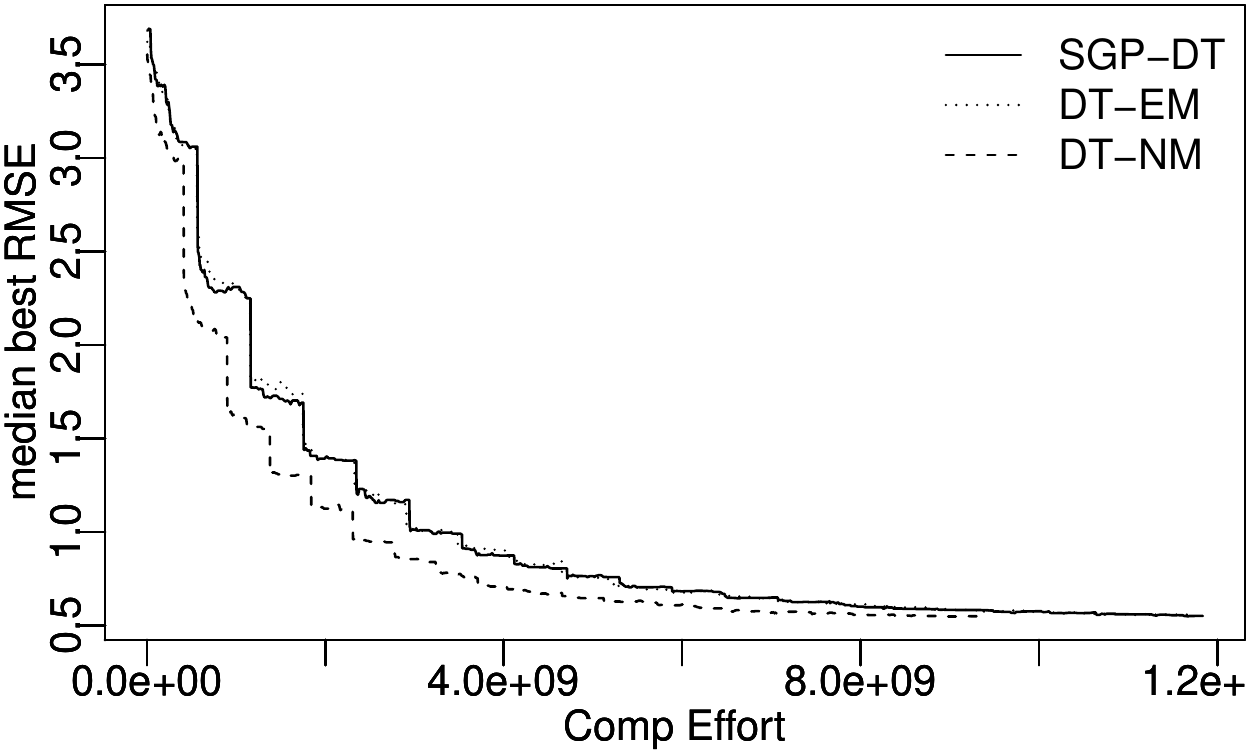}
		\vspace{\captionSpaceMedian}
		\caption{\textbf{enh}}\label{fig:RMSEmedian_enh}		
	\end{subfigure}	
\vspace{\spacev}
	\hspace{\hspaceMedian}
	\begin{subfigure}[t]{\zfMedian\linewidth}
		\centering
		\includegraphics[width=\linewidth]{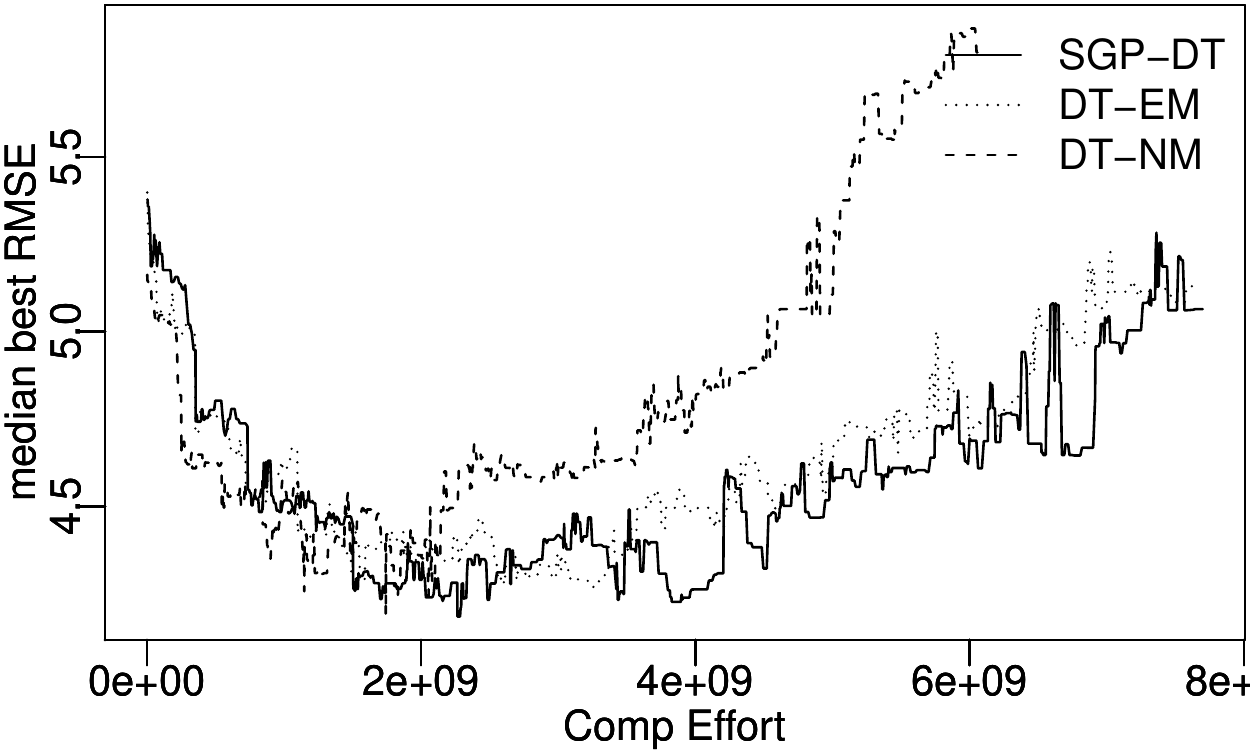}
		\vspace{\captionSpaceMedian}
		\caption{\textbf{housing}}\label{fig:RMSEmedian_housing}		
	\end{subfigure}
\vspace{\spacev}
	\hspace{\hspaceMedian}
	\begin{subfigure}[t]{\zfMedian\linewidth}
		\centering
		\includegraphics[width=\linewidth]{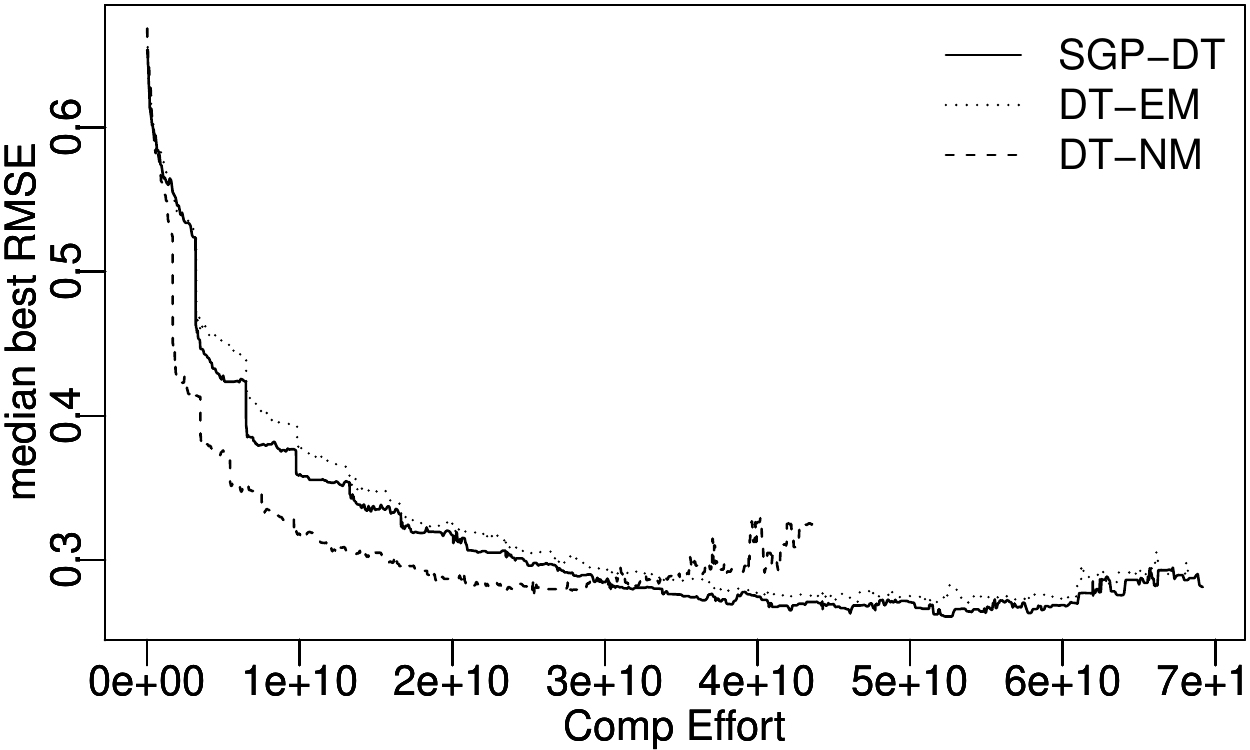}
		\vspace{\captionSpaceMedian}
		\caption{\textbf{tower}}\label{fig:RMSEmedian_tower}		
	\end{subfigure}
\vspace{\spacev}
	\hspace{\hspaceMedian}
	\begin{subfigure}[t]{\zfMedian\linewidth}
		\centering
		\includegraphics[width=\linewidth]{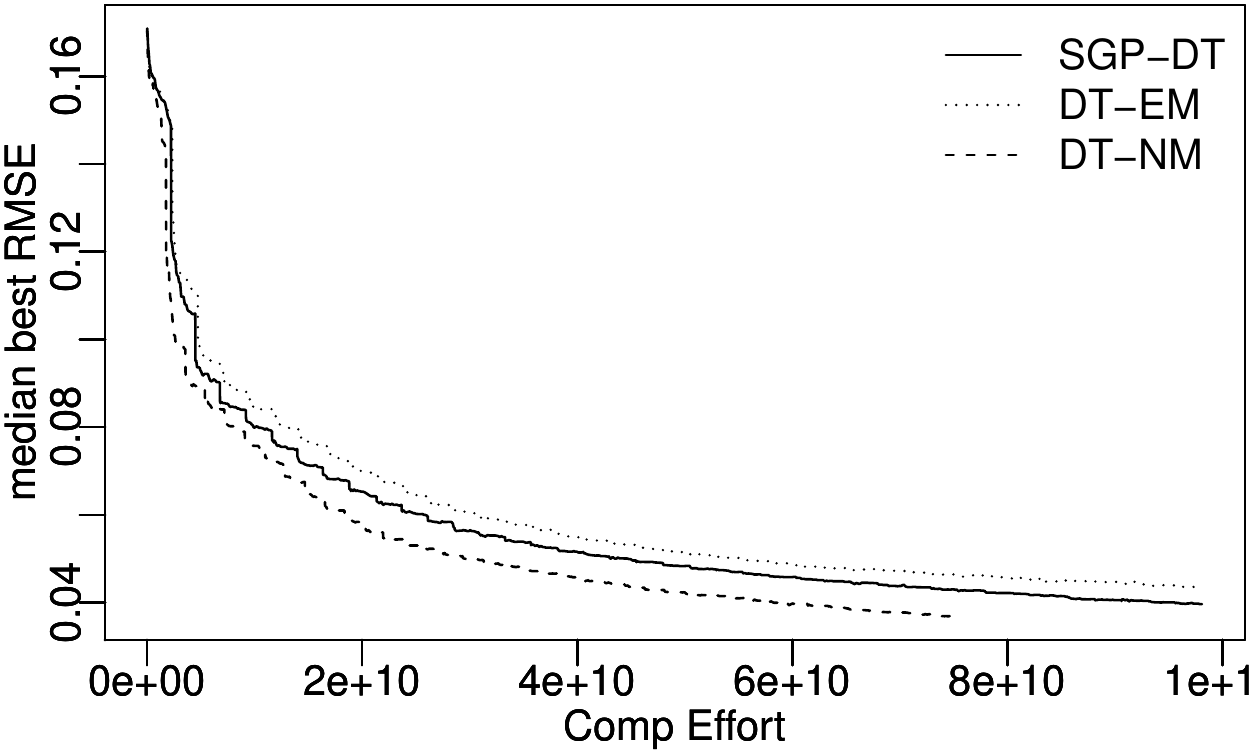}
		\vspace{\captionSpaceMedian}
		\caption{\textbf{uball5d}}\label{fig:RMSEmedian_uball5d}		
	\end{subfigure}
\vspace{\spacev}
	\hspace{\hspaceMedian}
	\begin{subfigure}[t]{\zfMedian\linewidth}
		\centering
		\includegraphics[width=\linewidth]{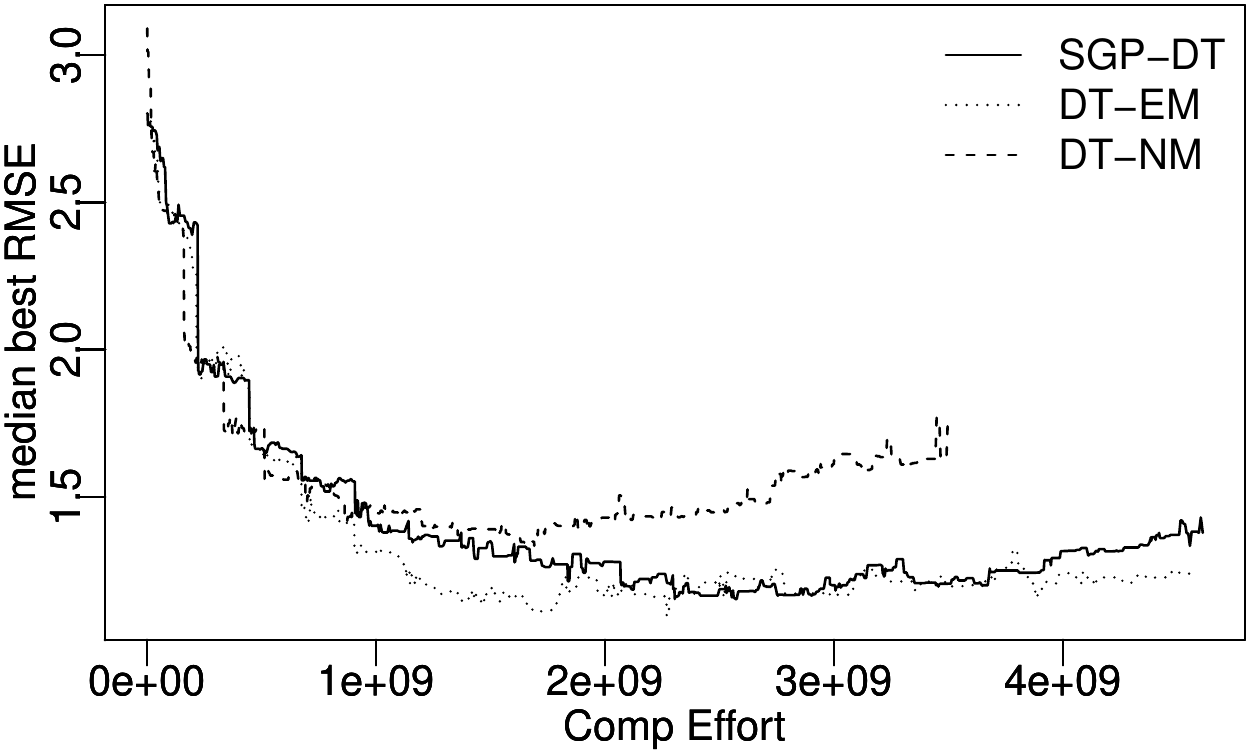}
		\vspace{\captionSpaceMedian}
		\caption{\textbf{yacht}}\label{fig:RMSEmedian_yacht}		
	\end{subfigure}	

	\caption{Median RMSE of the best so far on the test set by computational effort.}\label{fig:boxTestMediaEAll}
	
\end{figure*}

The eight plots indicate that \techSEES slightly overfits on the data sets \textit{tower} and \textit{yacht}, while on \textit{housing} produces a substantial overfitting, which is comparable to the one of \techSEESem but less severe than the one of \techSEESnm.
\techSEESem overfits four data sets: \textit{airfoil} (Fig.\ref{fig:RMSEmedian_airfoil}), \textit{housing} (Fig.\ref{fig:RMSEmedian_housing}), \textit{tower} (Fig.\ref{fig:RMSEmedian_tower}), \textit{yacht} (Fig.\ref{fig:RMSEmedian_yacht}). The worst performance is from \techSEESnm that shows severe overfitting on \textit{airfoil} (Fig.\ref{fig:RMSEmedian_airfoil}), \textit{housing} (Fig.\ref{fig:RMSEmedian_housing}), \textit{tower} (Fig.\ref{fig:RMSEmedian_tower}) and \textit{yacht} (Fig.\ref{fig:RMSEmedian_yacht}).
Note that all three techniques overfit for the data sets \textit{yacht} (Fig.\ref{fig:RMSEmedian_yacht}) and \textit{housing} (Fig.\ref{fig:RMSEmedian_housing}). This can be explain by their relatively low number of instances (see Table~\ref{table:dataset}).

For the data sets \textit{concrete} (Fig.\ref{fig:RMSEmedian_concrete}),  \textit{enc} (Fig.\ref{fig:RMSEmedian_enc}) and \textit{enh} (Fig.\ref{fig:RMSEmedian_enh}) all three techniques do not manifest overfitting  (yet). Interestingly, in these three cases \mbox{\tooln} arrives to a low RMSE with less computations than \tool and \mbox{\toole}.
We conjecture that this is because \textit{concrete}, \textit{enc} and \textit{enh} are problems that do not need the additional expressiveness of the \emph{Min} and \emph{Max} symbols.

\tooln is the technique that yields to the smallest individuals, as such we would expect less overfitting. 
Surprisingly, this is not the case.
We believe that, to compensate the absence of discontinuity that \textit{Max} and \textit{Min} introduce, \tooln 
used the protected divisions more frequently.
This may lead to many asymptotic discontinuities, which are known to increase the overfitting~\cite{keijzer2003LS}.

When considering each data set individually, \tool and \toole mostly manifest similar overfitting, while \tooln manifests overfitting much earlier.
This suggests that (i) the non-terminal symbols \emph{Max} and \emph{Min} help to alleviate the overfitting problem; and (ii) relying on the variance (\tool) rather than MSE (\toole) in the fitness function indeed contributes to reduce RMSE (5.39\% on average, see Table~\ref{table:rmse}) but not to influence overfitting.

\section{Conclusion}\label{sec:conclusion}

In this paper, we proposed \tool, a new evolutionary technique that dynamically discovers and resolves intermediate dynamic targets.
Our key intuition is that the synergy of the linear scaling and mutation helps to exchange good genetic materials during the evolution.
Notably, \tool does not rely on any form of crossover, and thus without suffering from its intrinsic limitations~\cite{pawlak_review_2015,poli_schema_1998}.
Our experimental results confirm our intuitions and show that \tool outperforms \lexicase in both lower RMSE and less computational cost.
This is a promising result as \lexicase outperforms many GP-inspired algorithms~\cite{bigcomparison}.

This paper sparks interesting future work:

We do not perform any type of post-processing of the final solutions to reduce their size.
Indeed, the solutions may contain redundant elements.
We are currently investigating a post-processing step to minimize the size of the final solutions.

A possible future research direction is to automatically identify the proper number of iterations of \tool. Indeed, problems with different complexity and nature may require a different number of external and internal iterations.

%% file: algo/algo.tex
\SetAlFnt{\scriptsize}

\begin{algorithm}[tb]

\setlength{\AlCapSkip}{1mm}
	\linespread{0.9}\selectfont

	\caption{\textbf{\techSEES} }
	\label{alg:search}
	\small 
\DontPrintSemicolon
\Input{%
	$\overline{x}$ : values of the independent variables of the training cases \newline
    \textit{$\hat{y}$} : values of the dependent variables of the training cases \newline
   \textit{$\text{N}_{\textit{ext}}$} : number of external iterations \newline
    \textit{$\text{N}_{\textit{int}}$} : number of internal iterations
}
\Output{%
	    \textit{finalModel} : final regression model
}
\BlankLine

$ \langle \overline{x}_\text{val}, \hat{y}_\text{val}  \rangle \gets \textsc{split}( \overline{x},\hat{y})$\; \label{algo:split-val-start}
$\overline{x} \gets \{\overline{x} \backslash \overline{x}_\text{val}\}$\;
$\hat{y} \gets \{\hat{y} \backslash \hat{y}_\text{val}  \}$\; \label{algo:split-val-stop}
$\text{target}  \gets \hat{y} $ \label{algo:init-target} \;
$\textit{models}\gets \emptyset $\label{algo:init-models}  \;

		\For{ $\text{ext-iter} $ $ 1 \dots N_{\text{ext}}$ \label{algo:for-ext-start} }{ 
					$\mathcal{P} \gets \textsc{get-random-initial-population()}$ \label{algo:random-pop} \;
					\For{ $\text{int-iter} $ $ 1 \dots N_{\text{int}}$ }{   \label{algo:int-iter}
						\For{ each $\mathcal{I} \in \mathcal{P}$ }{ 
					             $\mathcal{I}_\text{ls} \gets \textsc{compute-ls}(\mathcal{I},\overline{x},\textit{target})$\label{algo:ls}\tcp*{linear scaling} 
					                \textit{fitness}$(\mathcal{I}) \gets \sigma^2( \textit{sem}(\mathcal{I}_\text{ls}(\overline{x}))   -  \textit{target})$ \tcp*{$\sigma^2$ variance} \label{algo:fitness}
				             }
			            $\mathcal{I}^\star_{ls} \gets \textsc{get-best-individual}(\mathcal{P})$ \label{algo:best-ind}\;
			            
			             $\textit{error} \gets target - \text{sem}(\mathcal{I}^\star_{ls}(\overline{x}))$ \;
			             add  $\mathcal{I}^\star_{ls}$ to $models$\; \label{algo:collect-models}

			             $\mathcal{P}^\prime \gets \emptyset$\;
			             add \textsc{elite}($\mathcal{P}$) to $\mathcal{P}^\prime$ \label{algo:elite}\;
		  				\While{ $\mathcal{P}^\prime$ is not full }{  \label{algo:offspring-start}
		  							$\mathcal{I} \gets \textsc{tournament-selection}(\mathcal{P})$\;
		  							add \textsc{mutate}($\mathcal{I}$) to $\mathcal{P}^\prime$\; \label{algo:offspring-stop}
		  			    } 
	  			    $\mathcal{P} \gets \mathcal{P}^\prime$\;
	  			    
					 }               
			        	

			$target  \gets  error$ \label{algo:for-ext-end}   \label{algo:new-target}
\tcp*{update the target}
		}

		$\textit{bestModels} \gets \textsc{validate-and-select} (  \overline{x}_\text{val}, \hat{y}_\text{val}   \textit{, models})$ \tcp*{best MSE models on val} \label{algo:validation}
		$finalModel  \gets \sum_{\textit{model} \in \textit{bestModels} } \textit{model} $ \; \label{algo:linear-comb}
		\Return{ \textit{finalModel}} \;

\end{algorithm}

%% file: tables/subjects-new.tex
	\renewcommand\arraystretch{1.1}
	\setlength{\tabcolsep}{4pt}

	\resizebox{\linewidth}{!}{
		\rowcolors{1}{}{gray!10}
		
		\begin{tabular}{lrrc|lrrc}
			\hiderowcolors
			\toprule
			\textbf{name} & \textbf{\# attributes} & \textbf{\# instances} & \textbf{source} & \textbf{name} & \textbf{\# attributes} & \textbf{\# instances} & \textbf{source} \\
			\midrule
			\showrowcolors
			airfoil &5 &1,503 & \cellcolor{white} & 	housing &14&506& \cellcolor{white}\\
			concrete& 8 &1,030 & \cellcolor{white} & tower &25&3,135 &\cellcolor{white}  \\ 
			enc &8 &768 &\cellcolor{white} & yacht &6&309& \cellcolor{white} \multirow{-3}{*}{UCI~\cite{asuncion2007uci}}  \\

			enh & 8&768 & \cellcolor{white} \multirow{-4}{*}{UCI~\cite{asuncion2007uci}}  & 			uball5d &5&6,024&\cite{white2013better}\\
			\bottomrule
		\end{tabular}
	}

%% file: tables/rmse-new.tex
\renewcommand\arraystretch{1.3}
	\setlength{\tabcolsep}{2pt}
	\resizebox{\linewidth}{!}{
		\rowcolors{1}{}{gray!10}
\newcolumntype{R}[1]{>{\raggedleft\arraybackslash}p{#1}}
	\begin{tabular}{lR{20mm}R{20mm}R{20mm}R{20mm}R{20mm}lR{20mm}R{20mm}R{20mm}R{20mm}}
		\hiderowcolors
		\toprule
		\textbf{}        & \multicolumn{5}{c}{\textbf{Root Mean Square Error (RMSE)} }   && \multicolumn{4}{c}{\textbf{Median RMSE  \% decrease of \tool over:} }   \\ \cmidrule{2-6} \cmidrule{8-11}
 \multicolumn{1}{c}{\textbf{Data set}} & \textbf{\techSEES} &  \textbf{\lasso} & \textbf{\lexicase} & \textbf{\techSEESem} & \textbf{\techSEESnm}  & & \textbf{\lasso} & \textbf{\lexicase} & \textbf{\techSEESem} & \textbf{\techSEESnm} \\
\showrowcolors
\midrule
		
	airfoil  & 2.4634 & 4.8484  & 3.6505 & 2.5643 & 2.9237 &  & 49.19 \% & 32.52 \%& 3.94 \% & 15.75 \%\\
	concrete & 6.5123 & 10.5383 & 7.0707 & 6.4476 & 6.4132 &  & 38.20 \%& 7.90 \%  & -1.00 \% & -1.55 \% \\
	enc      & 1.4838 & 3.2498  & 1.8647 & 1.4993 & 1.4584 &  & 54.34 \% & 20.43 \% & 1.03 \%  & -1.75 \% \\
	enh      & 0.5560 & 2.9645  & 1.2952 & 0.5714 & 0.5410 &  & 81.25 \% & 57.07 \% & 2.70 \%  & -2.76 \% \\
	housing  & 4.4700 & 4.9155  & 4.2785 & 4.4377 & 4.5273 &  & 9.06 \%& -4.48 \%& -0.73 \%& 1.26 \%  \\
	tower    & 0.2606& 0.2953  & 0.2975 & 0.2900 & 0.2900 &  & 11.75 \%& 12.39 \%& 10.12 \%& 10.12 \%\\
	uball5d  & 0.0402 & 0.1939  & 0.0618 & 0.0430 & 0.0372 &  & 79.29 \%& 35.00 \%& 6.63 \%& -7.87 \%\\
	yacht    & 1.0221 & 9.0237  & 1.3577 & 1.2849 & 1.1786 &  & 88.67 \% & 24.72 \% & 20.45 \% & 13.28 \%\\
		\bottomrule
		&&&\multicolumn{4}{r}{\textbf{Average RMSE \% decrease:}} & \textbf{51.47 \%} &\textbf{23.19 \%}&\textbf{5.39 \%}&\textbf{3.31 \%} \\
				\bottomrule
	\end{tabular}
	}

%% file: images/rmse-plot-new.tex
\newcommand{\zfBoxTest}{0.206}
\newcommand{\captionSpace}{-40pt}

\begin{figure}[!t] 
\vspace{-3mm}
	\centering
	\begin{subfigure}[]{\zfBoxTest\linewidth}
		\includegraphics[width=\linewidth]{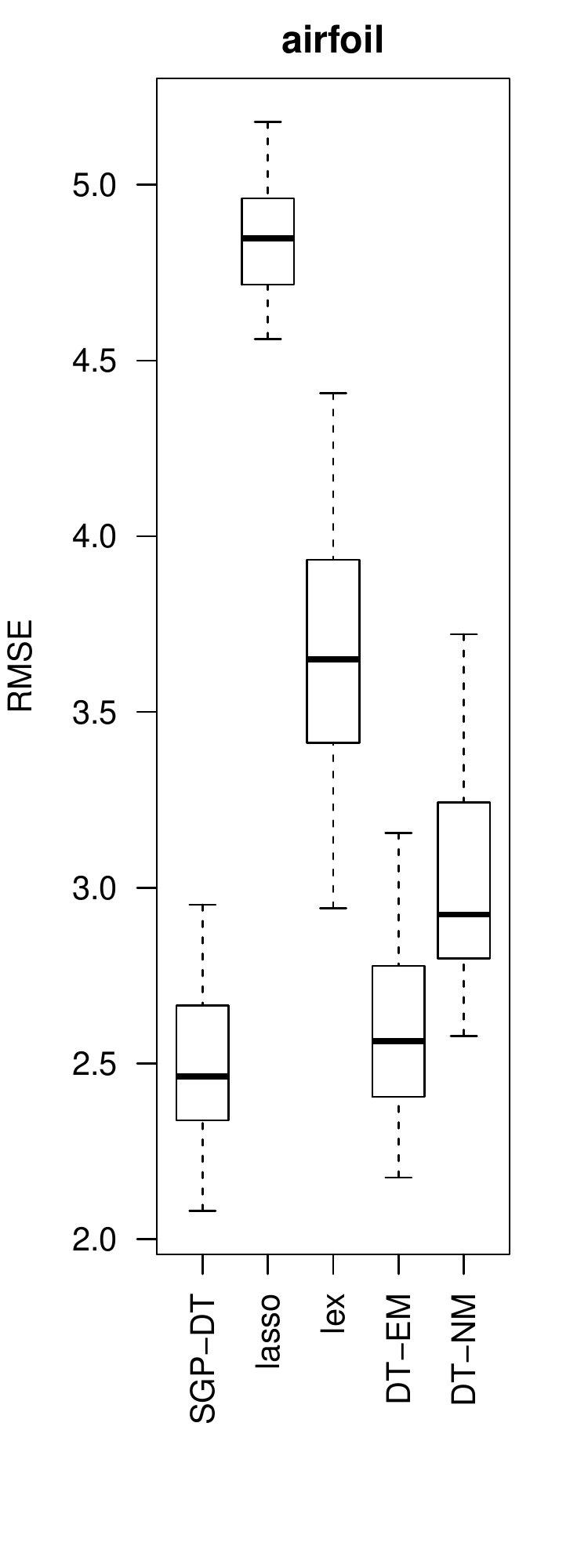}
		\vspace{\captionSpace}
		\caption{}\label{fig:RMSEbox_airfoil}		
	\end{subfigure}
	\quad
	\begin{subfigure}[]{\zfBoxTest\linewidth}
		\includegraphics[width=\linewidth]{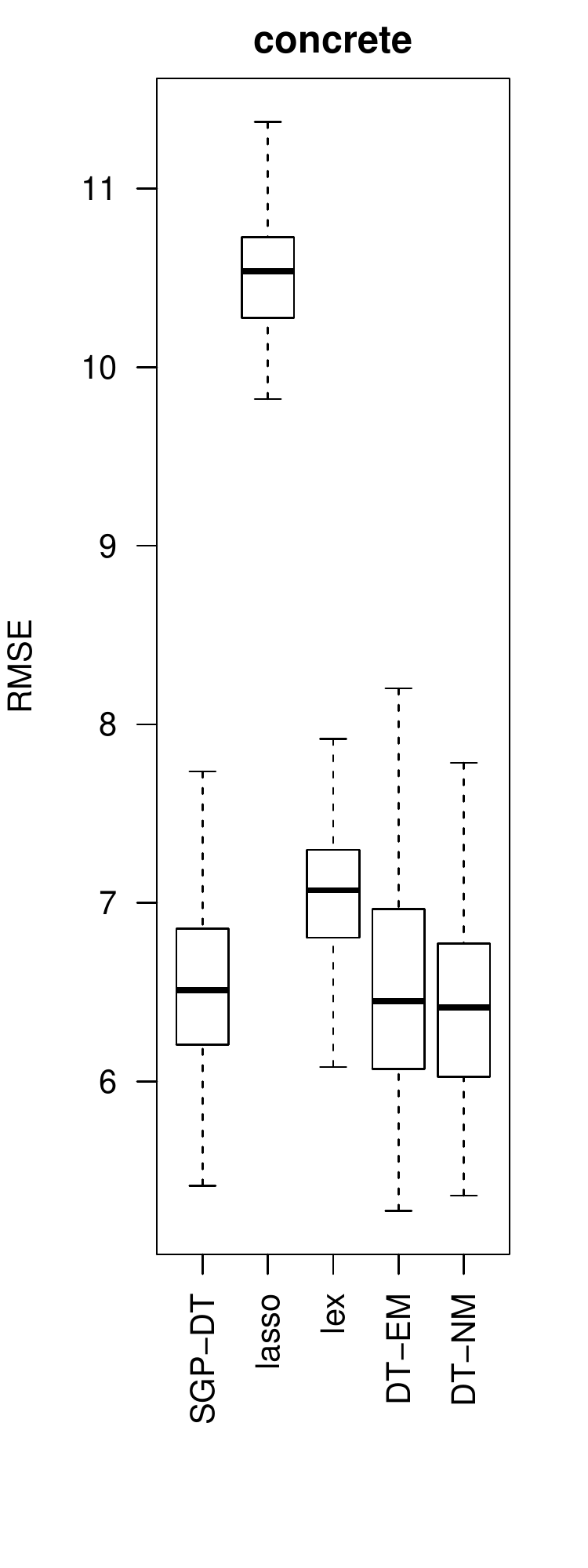}
		\vspace{\captionSpace}
		\caption{}\label{fig:RMSEbox_concrete}		
	\end{subfigure}
	\quad
	\begin{subfigure}[]{\zfBoxTest\linewidth}
		\includegraphics[width=\linewidth]{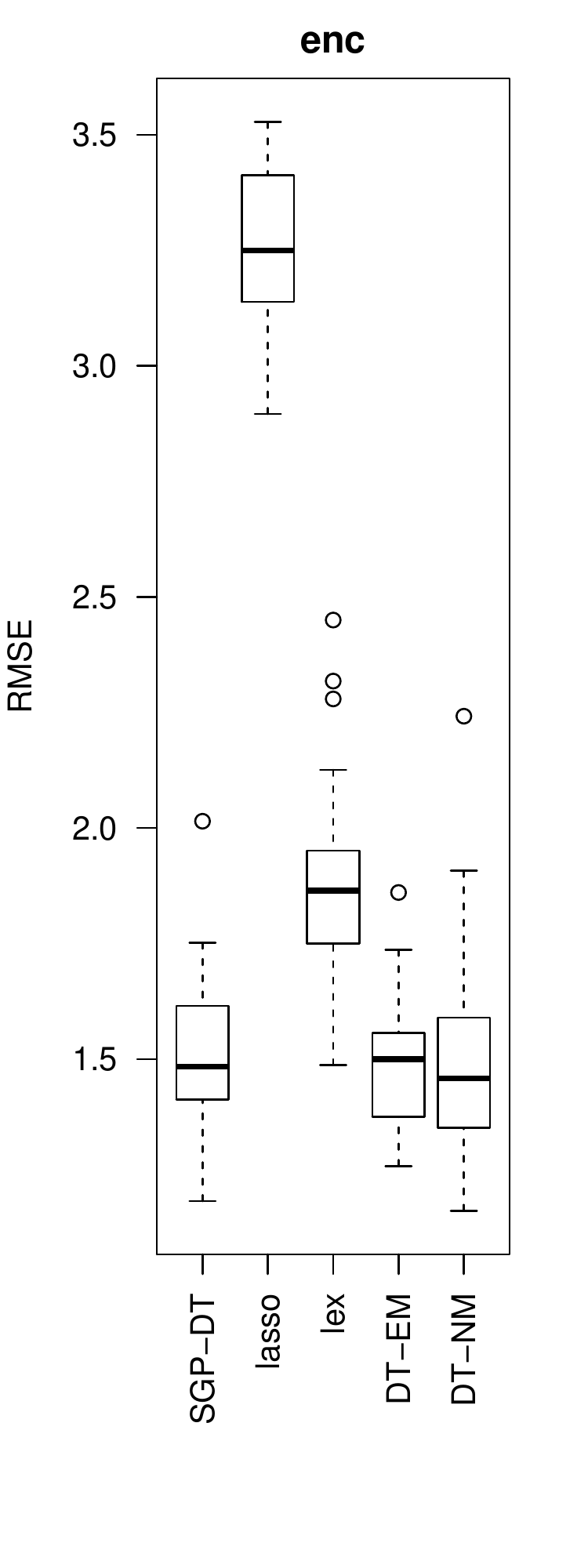}
		\vspace{\captionSpace}
		\caption{}\label{fig:RMSEbox_enc}		
	\end{subfigure}
	\quad	
	\begin{subfigure}[]{\zfBoxTest\linewidth}
		\includegraphics[width=\linewidth]{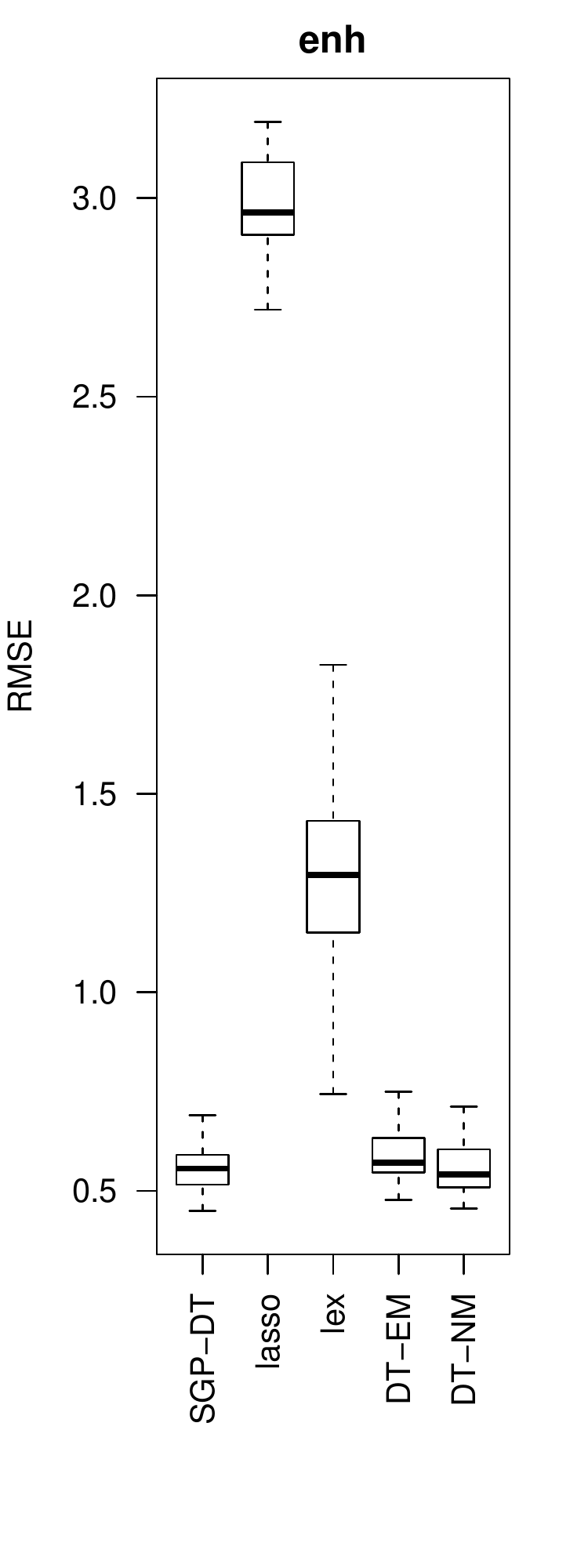}
		\vspace{\captionSpace}
		\caption{}\label{fig:RMSEbox_enh}		
	\end{subfigure}	
	\quad
	\begin{subfigure}[]{\zfBoxTest\linewidth}
		\includegraphics[width=\linewidth]{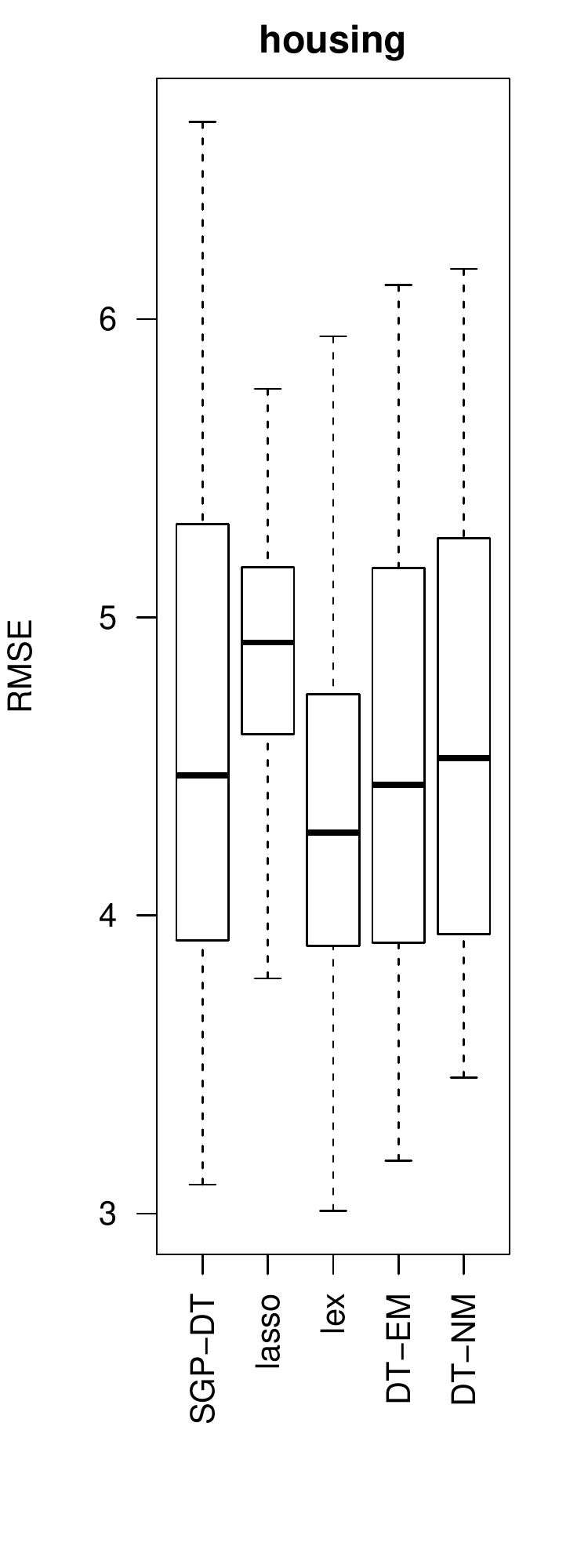}
		\vspace{\captionSpace}
		\caption{}\label{fig:RMSEbox_housing}		
	\end{subfigure}
	\quad
	\begin{subfigure}[]{\zfBoxTest\linewidth}
		\includegraphics[width=\linewidth]{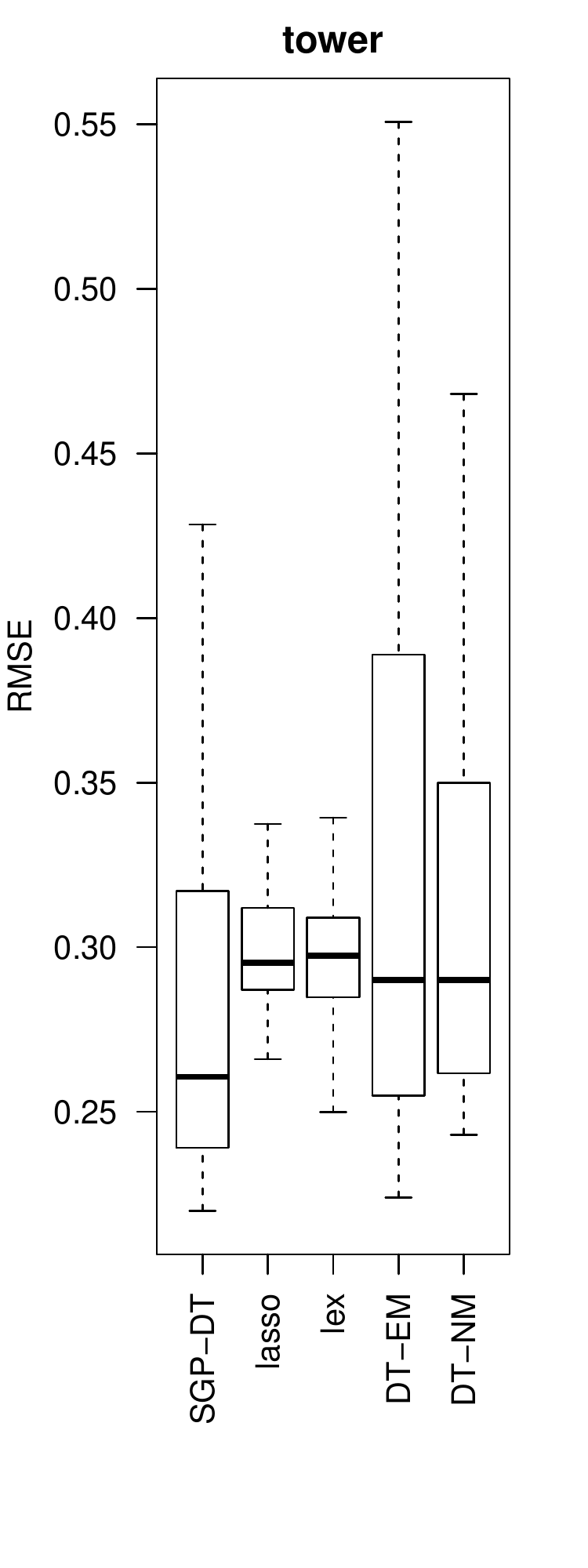}
		\vspace{\captionSpace}
		\caption{}\label{fig:RMSEbox_tower}		
	\end{subfigure}
	\quad
	\begin{subfigure}[]{\zfBoxTest\linewidth}
		\includegraphics[width=\linewidth]{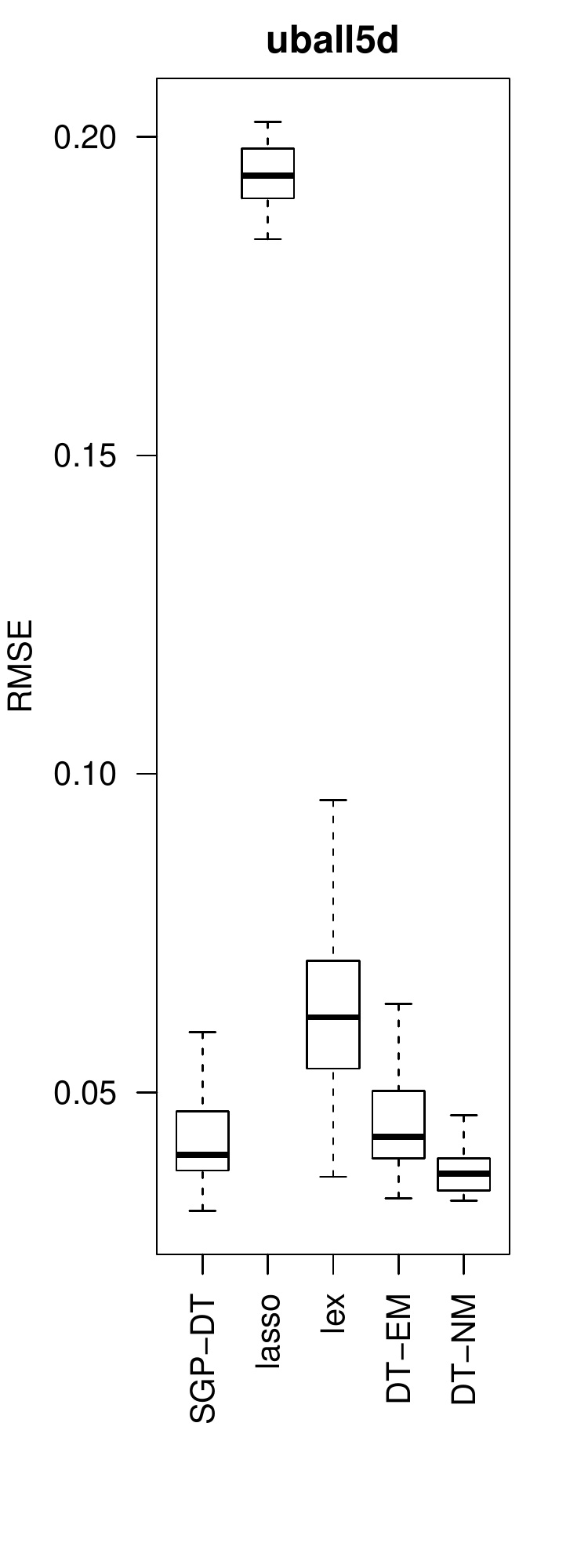}
		\vspace{\captionSpace}
		\caption{}\label{fig:RMSEbox_uball5d}		
	\end{subfigure}
	\quad
	\begin{subfigure}[]{\zfBoxTest\linewidth}
		\includegraphics[width=\linewidth]{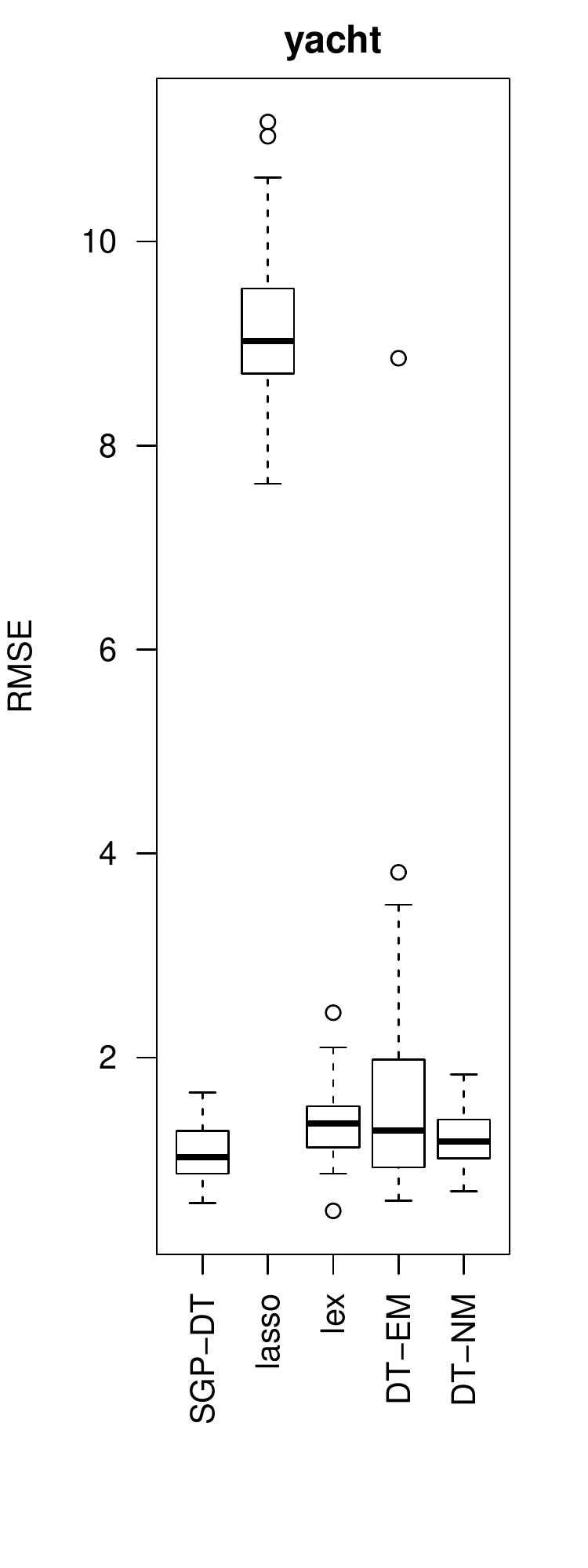}
		\vspace{\captionSpace}
		\caption{}\label{fig:RMSEbox_yacht}		
	\end{subfigure}	
			\vspace{-3mm}
	\caption{ RMSE of test set for all the techniques and for all the eight data sets. 
		\vspace{-1mm}
}\label{fig:boxTestRMSEAll}

\end{figure}

%% file: tables/nodes-new.tex
\renewcommand\arraystretch{1.2}
	\setlength{\tabcolsep}{4pt}
	\resizebox{\linewidth}{!}{
		\rowcolors{1}{}{gray!10}
		\newcolumntype{R}[1]{>{\raggedleft\arraybackslash}p{#1}}
	\begin{tabular}{lR{20mm}R{28mm}R{28mm}R{28mm}lR{20mm}R{20mm}R{20mm}}
		\hiderowcolors
		\toprule
		\textbf{}        & \multicolumn{4}{c}{\textbf{Median number of evaluated nodes}}         &  & \multicolumn{3}{c}{\textbf{Reduction ratio of \techSEES over}}                                        \\ \cmidrule{2-5} \cmidrule{7-9}
\textbf{Data set} & \textbf{\techSEES} & \textbf{\lexicase} & \textbf{\techSEESem} & \textbf{\techSEESnm}  &        & \textbf{\lexicase} & \textbf{\techSEESem} & \textbf{\techSEESnm}    \\
\showrowcolors
\midrule
		airfoil          & 1.00E+10           & 9.28E+10           & 1.00E+10             & 9.03E+09              &        & 9.26$\times$                                 & 1.00$\times$                   & 0.90$\times$    \\
		concrete         & 1.14E+10           & 6.43E+10           & 1.14E+10             & 8.82E+09             &         & 5.64$\times$                                 & 1.00$\times$                   & 0.77$\times$    \\
		enc              & 1.18E+10           & 4.99E+10           & 1.17E+10             & 9.37E+09               &       & 4.25$\times$                                 & 0.99$\times$                   & 0.80$\times$    \\
		enh              & 1.18E+10           & 5.08E+10           & 1.17E+10             & 9.27E+09            &          & 4.30$\times$                                 & 0.99$\times$                   & 0.78$\times$    \\
		housing          & 7.70E+09           & 3.09E+10           & 7.63E+09             & 6.03E+09             &         & 4.02$\times$                                 & 0.99$\times$                   & 0.78$\times$    \\
		tower            & 7.21E+10           & 1.94E+11           & 7.12E+10             & 4.45E+10           &           & 2.69$\times$                                 & 0.99$\times$                   & 0.62$\times$    \\
		uball5d          & 9.83E+10           & 3.94E+11           & 9.76E+10             & 7.50E+10            &          & \multicolumn{1}{r}{4.01$\times$}               & \multicolumn{1}{r}{0.99$\times$} & 0.76$\times$   \\
		yacht            & 4.62E+09           & 2.00E+10           & 4.58E+09             & 3.47E+09          &            & \multicolumn{1}{r}{4.34$\times$}               & \multicolumn{1}{r}{0.99$\times$} & 0.75$\times$    \\
		\bottomrule
			\multicolumn{6}{r}{\textbf{Average reduction ratio:}} &\textbf{4.81$\times$}&\textbf{0.99$\times$}&\textbf{0.77$\times$}\\
		\bottomrule
	\end{tabular}
	}